\documentclass[preprint,12pt,authoryear]{elsarticle}

\usepackage{amssymb}
\usepackage{amsmath}
\usepackage{comment}
\usepackage{amsmath,amssymb,amsfonts}
\usepackage{graphicx}
\usepackage{textcomp}
\usepackage[table]{xcolor}
\usepackage{subcaption}
\usepackage{hyperref}
\usepackage{balance}
\usepackage{dblfloatfix}
\usepackage[font={small,it}]{caption}
\usepackage{tabularx}
\usepackage{float}
\usepackage{svg}
\usepackage{makecell}
\usepackage{enumitem}
\usepackage{pifont}
\usepackage{multirow}
\usepackage{booktabs}
\usepackage{arydshln}
\usepackage{colortbl}
\usepackage{diagbox}
\usepackage{comment}
\usepackage{dsfont}
\usepackage[ruled,vlined]{algorithm2e}
\usepackage[cache=false]{minted}
\usepackage{microtype}

\journal{Nuclear Physics B}

\begin{document}

\begin{frontmatter}


\author{Mohammad Meymani}
\ead{mohammad.meymani79@unb.ca}
\author{Roozbeh Razavi-Far}
\ead{roozbeh.razavi-far@unb.ca}
\affiliation{organization={Trustworthy and Secure AI (TSAI) Lab, Faculty of Computer Science, University of New Brunswick},
city={Fredericton}, country={Canada}}
\title{{KAN-Robust-Bench: A Benchmark for Evaluating the Robustness of Kolmogorov-Arnold Networks}} 
\begin{abstract}
While machine learning models have demonstrated strong performance in many domains, these models have shown profound vulnerabilities when they are exposed to adversarial threats. While adversarial attacks fall into various categories, the most prominent category in research studies is evasion. In evasion attacks, the adversary generates perturbed versions of samples, which might not be observable by human eyes. These samples generally fool the machine learning models with high confidence. This phenomenon poses a significant security violation against machine learning models. In this paper, we investigate the certified and empirical robustness of various Kolmogorov-Arnold network architectures against strong evasion attacks. At first, we provide the mathematical foundations for randomized smoothing and interval bound propagation, and report the $\ell_2$-certified robustness of the models under randomized smoothing. After that, we systematically evaluate the robustness of various defended and undefended KAN models under FGSM, PGD, and C\&W attacks in order to find out the optimal defense strategies and architectures.
\end{abstract}


\end{frontmatter}

\section{Introduction}
{\underline{M}achine \underline{l}earning (ML) has extended \underline{a}rtificial \underline{i}ntelligence (AI) by moving systems from hard-coded rules to data-driven learning. ML has facilitated numerous tasks in a huge number of fields including retail, finance, \underline{n}atural \underline{l}anguage \underline{p}rocessing (NLP), autonomous driving, \underline{c}omputer \underline{v}ision (CV), healthcare, cybersecurity, and many other domains \cite{shinde2018review,sarker2021machine}.} ML models have evolved during decades from classical networks and \underline{m}ulti-\underline{l}ayer \underline{p}erceptron (MLP) to deep neural networks. Each learning paradigm aims to solve the problems that are infeasible, hard, or inefficient for the previous ones \cite{boutaba2018comprehensive}. \underline{K}olmogorov-\underline{A}rnold \underline{n}etworks (KANs) have been introduced to substitute MLP models in numerous areas \cite{liu2025kan,ji2024comprehensive,somvanshi2025survey}. 

\underline{K}olmogorov-\underline{A}rnold representation \underline{t}heorem (KAT) states that any continuous multivariate function $f$ can be written as a composition of finite number of continuous univariate functions. This can be formalized as follows:
\begin{equation}
    f(x_1,~x_2,~...,~x_n) = \sum_{j=1}^m \psi_j(\sum_{i=1}^n \phi_{ij}(x_i)),
\end{equation}
where $n$ is the number of inputs, $m$ is the number of compositions,  $x_i$ is the $i^{th}$ input, $\phi_{ij}$ is the univariate function for $x_i$ in $j_{th}$ composition, and $\psi_j$ is the $j^{th}$ composition. KANs make use of KAT to provide powerful alternatives to MLP models \cite{ji2024comprehensive,somvanshi2025survey,lou2026kolmogorov}.

\begin{figure}[h]
    \centering
    \includegraphics[width=0.5\linewidth]{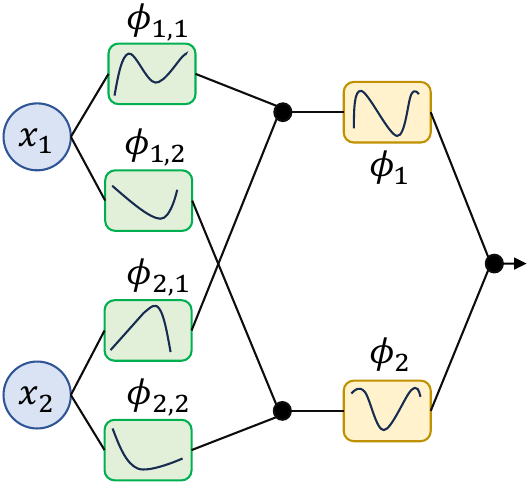}
    \caption{A simple KAN architecture consisting of four Splines at the first hidden layer and two Splines at the second hidden layer.}
    \label{fig:KAN-Architecture}
\end{figure}

KANs and MLPs differ in multiple aspects. First, KANs can outperform MLPs on specific tasks, including time-series analysis, with fewer trainable parameters. Second, in general, the training time for a KAN is significantly slower than that of an MLP on the same task \cite{ji2024comprehensive,somvanshi2025survey}. Third, KANs exhibit greater robustness to catastrophic forgetting in specific low-dimensional tasks than MLPs, making them suitable options for continual learning \cite{cacciatore2024preliminary,rahman2026catastrophic}. Figure \ref{fig:KAN-Architecture} illustrates a simple architecture for a KAN, where traditional activation functions in MLPs are replaced by learnable ones. After each hidden layer, every connected Spline is summed together and sent to the next layer.

\underline{A}dversarial \underline{m}achine \underline{l}earning (AML) is an interdisciplinary field, studying the vulnerabilities of ML models. Adversarial attacks exploit these vulnerabilities to harm utility, privacy, or explainability of ML models. Adversarial defenses, on the other hand, aim to secure ML models during different stages of ML lifecycle. This field faces an evolving arm-race, where new attacks and defenses are introduced periodically. This necessitates the researchers to consider designing built-in defense systems when developing ML models \cite{vashagh2026recent,kurakin2016adversarial,meymani2026divided}.

Adversarial attacks are divided into three major categories: evasion, poisoning, and exploratory attacks. In evasion attacks, the attacker generates adversarial samples by adding subtle and carefully crafted noise to the original data. These adversarial samples are imperceptible to human observers, but can fool the ML model with high confidence. Poisoning attacks aim to corrupt the training data by tampering with features and/or labels to harm the training process. Finally, exploratory attacks aim to violate the privacy of ML models by inferring their parameters and/or training data \cite{vashagh2026recent,meymani2026defense}.

Moreover, an adversary can possess different amounts of knowledge about the target model's sensitive information including gradients, training data, and parameters. If an adversary has perfect knowledge about the model, it is considered as a white-box adversary. On the other hand, an adversary with zero information is considered as a black-box one. Finally, a gray-box adversary sits between white-box and black-box, where partial knowledge about the target model is available \cite{vashagh2026recent,meymani2026defense}. 

Certified robustness in AML refers to a formal and mathematically provable guarantees that a model’s prediction will remain unchanged within a specified perturbation bound and norm. There are various techniques that provide certified robustness by analyzing
the robustness from a different perspective. These techniques include \underline{r}andomized \underline{s}moothing (RS), \underline{i}nterval \underline{b}ound \underline{p}ropagation (IBP), convex relaxation, verification, and \underline{s}emi\underline{d}efinite \underline{p}rogramming (SDP) relaxation \cite{cohen2019certified,gowal2018effectiveness,meng2022adversarial,zhang2020tightness,salman2019convex,raghunathan2018semidefinite}.

{ In this paper, we focus on white-box evasion attacks since they provide a direct means of robustness evaluation of the trained models during inference. This allows us to study the vulnerability of the learnt models when they are exposed to perturbed inputs. Moreover, evasion attacks are among the most widely studied adversarial threats, making them a suitable basis for systematically comparing the robustness of different KAN architectures and defense strategies.} Our contributions are as follows:
\begin{itemize}
    \item We evaluate the empirical robustness of state-of-the-art KAN-based computer vision architectures against strong white-box evasion attacks.
    \item We systematically evaluate the impact of adversarial training, randomized smoothing, and interval bound propagation on the robustness of KAN architectures across CIFAR-10 and SVHN.
    \item We analyze the certified robustness under randomized smoothing, and report certified accuracy and certified radius across different KAN architectures and perturbation bounds.
\end{itemize}

The remainder of the paper is as follows. In Section \ref{sec:related-works}, we introduce the related research studies. In Section \ref{sec:problem-setting}, we discuss the problem settings. In Section \ref{sec:methodology}, we explain our methodology by providing defense algorithms and mathematical underpinning for RS and IBP. In Section \ref{sec:experimental-setup-results}, we discuss experimental setups, and  demonstrate the certified and empirical results. Finally, in Section \ref{sec:conclusion}, we suggest future research directions and conclude our work.
\section{Related Works}
\label{sec:related-works}
KANs are used for a variety of regression and classification tasks, including time series analysis, graph learning, signal processing healthcare, and CV. For CV, architectures such as KAN-Mixers \cite{canuto2025kan}, KANICE \cite{ferdaus2024kanice}, and PoolKANNeXt \cite{lv2025poolkannext} have been proposed.

KAN-Mixers \cite{canuto2025kan} extends MLP-mixer \cite{tolstikhin2021mlp} architecture by replacing the standard MLP-based layers with KAN layers. Similar to MLP-Mixer, KAN-Mixers performs token- and channel-mixing to learn both spatial and feature-level relationships. KAN-Mixers consists of a patch-embedding layer, mixer blocks, an adaptive average pooling layer, and a classification head. The results indicated that KAN-Mixers achieved the highest accuracy on Fashion MNIST and CIFAR-10 datasets compared to a standard KAN, MLP, and MLP-Mixer.

KANICE \cite{ferdaus2024kanice} is a hybrid architecture that combines KAN and CNN. KANICE consists of \underline{I}nteractive \underline{C}onvolution \underline{B}lock (ICB), standard convolutional layers, batch normalization, pooling layers, and KAN linear layers. ICBs are responsible for capturing spatial relationships, serving as initial feature extractor. Standard convolutional layers are used after each ICB to further process and extract the features. Batch normalization is used to stabilize the learning process, and pooling layers are used to reduce the dimensionality. Finally KAN linear layers are used to approximate complex functions more accurately. The results on MNIST, Fashion MNIST, EMNIST, and SVHN demonstrated the superior performance of KANICE compared to CNN, CNN\_KAN, ICB\_CNN, and ICB\_KAN architectures.

PoolKANNeXt \cite{lv2025poolkannext} is another hybrid KAN-CNN architecture, which combines pooling-based feature-mixing, dual-activation KAN blocks, and ConvNeXt-style \cite{liu2022convnet} hierarchical architecture. PoolKANNeXt consists of four stages: stem, main, downsampling, and output. The stem phase is the initial feature extractor. The main phase consists of average pooling, convolution blocks, and GELU and Swish activation functions. After that, the inputs from previous phases are downsampled and sent to the output layer. While this model was originally designed for biomedical image classification, it demonstrated strong performance in general CV-based classification tasks including CIFAR-10.

In \cite{ostanin2024evaluating}, the authors evaluated the robustness of a KAN model and a MLP model under Gaussian noise and adversarial attacks using MNIST dataset. The authors used \underline{F}ast \underline{G}radient \underline{S}ign \underline{M}ethod (FGSM) and \underline{P}rojected \underline{G}radient \underline{D}escent (PGD) to generate adversarial samples. While KAN achieved slightly higher accuracy than MLP on clean data, its performance decreased more sharply under noise and attacks. MLP model consistently outperformed the KAN model in terms of accuracy, precision, recall, and F1-score under noise and attacks. This trend indicated greater adversarial vulnerability of KAN compared to MLP.

In \cite{ibrahum2024resilient}, the authors compared the performance of some KAN-based architectures (i.e., KAN-Mixer and convolutional KAN) and MLP-based architectures (i.e., MLP-Mixer) under \underline{b}asic \underline{i}terative \underline{m}ethod (BIM), FGSM, and \underline{C}arlini and \underline{W}agner (C\&W). Experiments were conducted on BTSD, CTSD, and GTSRB datasets. Moreover, the authors investigated \underline{a}dversarial \underline{t}raining (AT) and randomized smoothing as defense strategies to further study the robustness. Results indicated that robustness depended on the architecture. KAN-Mixer generally outperformed MLP-Mixer in terms of robustness, while standard KAN generally demonstrated less robustness than standard MLP model. Randomized smoothing and adversarial training improved robustness across models, with randomized smoothing often providing the stronger defense.

In \cite{djosic2024kan}, the authors evaluated the robustness of an MLP and four KAN variants (linear KAN, Fourier KAN, Chebyshev KAN, and Jacobi KAN) under Gaussian noise, FGSM, and PGD attacks with MNIST as the dataset. MLP and linear KAN demonstrated the highest accuracies. Under noise MLP showed the greatest robustness, with linear KAN being the second best, while other KAN variants showed severe degradation. After applying FGSM, MLP showed the greatest robustness followed by the linear KAN. In contrast, Chebyshev KAN and Fourier KAN demonstrated greatest robustness under PGD, while MLP and linear KAN had the lowest scores.

In \cite{dong2024kolmogorov}, the authors analyzed the robustness of KANs for time series classification under PGD attack. The authors used UCR2018 dataset. The results indicated that KAN demonstrated similar or sometimes slightly better than MLP in terms of clean accuracy. Moreover, the KAN model showed greater adversarial robustness than the MLP model. As a result, the authors concluded that the greater robustness of the KAN model stems from its lower Lipschitz constant. 

In \cite{alter2024robustness}, the authors evaluated the robustness of fully connected convolutional KANs across a wide range of datasets and adversarial attacks, and compared them with \underline{c}onvolutional \underline{n}eural \underline{n}etworks (CNNs) and \underline{f}ully \underline{c}onnected \underline{n}eural \underline{n}etworks (FCNNs). The datasets included MNIST, Fashion MNIST, KMNIST, CIFAR-10, SVHN, and ImageNet. The attacks included both white-box (i.e., FGSM and PGD) and black-box (i.e., Square Attack) attacks. The results suggested that convolutional KAN models generally exhibited greater robustness compared to CNN models with comparable sizes. Moreover, the impact of spline order and number of knots depends on model scale; higher spline orders often improve robustness in small and medium models, whereas lower spline orders can be more robust in large models. The authors claimed that optimizing the hyperparameters of KAN-based architectures could be a promising research direction for robustness studies.

In \cite{schumacher2025training}, the authors evaluated the robustness of KAN, MLP, and CNN models under FGSM and PGD attacks on Fashion MNIST and CIFAR 10 datasets. The authors proposed GloroKAN to provide certified robustness, however, adversarially trained KAN outperformed their proposed approach.

{Although previous studies have investigated the adversarial robustness of KANs, most of them primarily focused on comparisons between KAN-based models and conventional architectures such as MLPs or CNNs. In contrast, our study focuses on robustness within the KAN family, aiming to determine how different KAN-based architectures behave under the same adversarial conditions. We evaluate KAN-Mixers, KANICE, and PoolKANNeXt across CIFAR-10 and SVHN, considering multiple defense strategies and a wider range of attack configurations. We further compare adversarial training, randomized smoothing, and interval bound propagation, while complementing the empirical evaluation with certified robustness analysis. This enables a focused evaluation of how architectural and defense choices influence robustness among KAN-based models.}
\section{Problem Settings}
\label{sec:problem-setting}
There are a variety of attacks that use white-box knowledge to exploit the vulnerabilities of machine learning models. These attacks use the model's parameters to generate adversarial samples, for instance, FGSM \cite{goodfellow2015explaining} is a one-step attack, generating adversarial samples by moving toward the gradient of the loss function:
\begin{equation}
x_{\text{adv}} = x + \epsilon \cdot \text{sign}(\nabla_x L(\theta, x, y)), 
\label{fgsm-formula}
\end{equation}
where $x_{adv}$ is the generated adversarial sample, 
$x$ is a benign sample, $y$ is the corresponding label of that sample, $\epsilon$ is the perturbation size, $L(\theta, x, y)$ is the loss function, $\nabla_x {L}$ is the gradient of loss function with respect to $x$, and $\theta$ denotes the model's parameters.

Another famous attack is PGD \cite{madry2017towards}, which is a multi-step attack unlike FGSM. PGD refines a random perturbation over iterations to find a desired perturbation causing misclassification. Eq. (\ref{PGD-Formula}) shows how an adversarial sample is generated using PGD:
\begin{equation}
x_{adv}^{i+1} = \prod_{x + \delta} \left(x_{adv}^i + \alpha \cdot sign \left( \nabla_x L(\theta, x, y) \right) \right),
\label{PGD-Formula}
\end{equation}
where $\delta$ is the perturbation, $\alpha$ is step size, $i$ is iteration number, $x_{adv}^i$ is the adversarial sample during $i^{th}$ iteration, and $\prod_{x + \delta}$ is the projection function, ensuring the input stays within perturbation bounds ($[x-\delta,~x+\delta]$).

Additionally, C\&W \cite{carlini2017towards} uses the following objective function:
\begin{equation}
\left\{
\begin{aligned}
&\text{minimize} \quad D(x, x+\delta) + \lambda.f(x+\delta)\\
&\text{subject to} \quad x+\delta \in[0,1]^n 
\end{aligned}
\right.
\end{equation}
where $D(x, x+\delta)$ is the distance between the benign and adversarial samples, {$n$ is the number of features (pixels)}, $\lambda$ controls the trade-off between perturbation size and attack success rate, $f(x+\delta)$ is an objective function, ensuring that $x+\delta$ is misclassified by the model.

Adversarial training is a defense strategy that aims to learn representations of both benign samples and adversarial samples during the training session. This can either be done by augmenting adversarial samples during the training session, or defining new objective function that learns both representations. Although this strategy has been widely used, most of the methods face robustness-accuracy trade-off. Robustness-accuracy trade-off occurs when clean accuracy is sacrificed to some extent to increase the robustness of the model \cite{meymani2026divided,xu2021robust,nath2023enhancing}.

Randomized smoothing generates a smoothed model by adding Gaussian noise to the inputs and and aggregating predictions via Monte Carlo sampling. In Monte Carlo voting, multiple noisy versions of the input are generated and the predictions are aggregated via majority voting \cite{cohen2019certified,nandi2023certified}.

Interval bound propagation provides a certified defense strategy by propagating the input perturbation bounds through the whole network. During IBP-based training, the training loss equals to the linear combination of clean loss and IBP loss to control the trade-off between accuracy and robustness. The final loss $\ell_\text{Final}$ can be formulated as follows:
\begin{equation}
    \ell_\text{Final} = (1-\lambda)\times \ell_\text{Clean}+\lambda\times \ell_\text{IBP},
\label{formula:IBP-Clean-Loss-Tradeoff}
\end{equation}
where $\ell_\text{Clean}$ is the standard loss function, $\ell_\text{IBP}$ is the IBP-based loss functions, and $\lambda$ controls the trade-off between robustness and accuracy \cite{gowal2018effectiveness,mao2024understanding,mirman2018differentiable}.

\section{Methodology}
\label{sec:methodology}
\begin{figure*}[h]
    \centering
    \includegraphics[width=\linewidth]{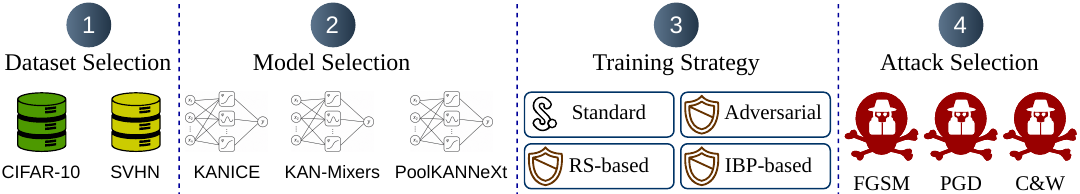}
    \caption{Step-by-step process of experimental methodology.}
    \label{fig:methodology}
\end{figure*}

In this section, we formulate certified robustness, explain defense mechanisms, and formalize the threat model. Figure \ref{fig:methodology} shows an step-by-step methodology process, where we begin by choosing a dataset, an architecture, and a training strategy to evaluate the performance of the models against a diverse set of adversarial attacks.

\subsection{Certified Robustness Formulation}
Unlike empirical defenses that rely on specific adversarial attacks and parameters, certified defenses provide mathematical guarantees on model behavior within predefined perturbation bounds \cite{cohen2019certified,nandi2023certified}.
\subsubsection{Randomized Smoothing}
Let $f$ denote a base KAN classifier. Randomized smoothing constructs a smoothed classifier by injecting Gaussian noise into the input space:

\begin{equation}
\eta \sim \mathcal{N}(0,\sigma^2 I),
\end{equation}
where $\sigma$ denotes the noise level and $I$ is the identity matrix \cite{cohen2019certified}. The smoothed classifier is defined as:

\begin{equation}
g(x)=\arg\max_{c\in\{1,\ldots,C\}}
P\left(f(x+\eta)=c\right),
\end{equation}
where $C$ is the number of classes. Let $p_A=P\left(f(x+\eta)=c_A\right)$ denote the probability of the most probable class $c_A$, and $p_B=\max_{c\neq c_A}P\left(f(x+\eta)=c\right)$ represent the probability of the second most probable class. According to the randomized smoothing certification theorem, the classifier prediction remains unchanged within an $\ell_2$-ball of radius:

\begin{equation}
R=
\frac{\sigma}{2}
\left(
\Phi^{-1}(p_A)
-
\Phi^{-1}(p_B)
\right),
\end{equation}
where $\Phi^{-1}(\cdot)$ is the inverse cumulative distribution function of the standard normal distribution \cite{cohen2019certified}. Therefore, for any perturbation $\delta$ satisfying $\|\delta\|_2 < R$, the prediction is guaranteed to remain unchanged:

\begin{equation}
g(x+\delta)=g(x).
\end{equation}

During inference, class probabilities ($\hat p_c$) are estimated using Monte Carlo sampling:

\begin{equation}
\hat p_c=
\frac{1}{M}
\sum_{i=1}^{M}
\mathds{1}
\bigl(
f(x+\eta_i)=c
\bigr),
\end{equation}
where $M$ is the number of noisy samples, $\mathds{1}$ is an indicator function that outputs one for true conditions, and outputs zero for false ones, and $\eta_i\sim\mathcal N(0,\sigma^2I)$. The certified accuracy of randomized smoothing is computed as:

\begin{equation}
CA_{RS}(r)=
\frac{1}{N}
\sum_{i=1}^{N}
\mathds{1}
\left(
R_i \ge r ~ \land ~ y_i=\hat{y}_i
\right),
\end{equation}
where $R_i$ is the certified $\ell_2$ radius of sample $i$, $N$ is the number of test samples, $r$ is the threshold, $y_i$ is the actual class label, and $\hat{y}_i$ is the predicted label. 
\subsubsection{Interval Bound Propagation}
Interval Bound Propagation certifies robustness by propagating perturbation intervals through every layer of the network \cite{gowal2018effectiveness,mao2024understanding}. Given an input sample $x$, the adversarial region is defined as:
\begin{equation}
\mathcal X=
\left\{
x+\delta:
\|\delta\|_\infty\le\epsilon
\right\},
\end{equation}
where $\epsilon$ is the perturbation bound. The initial interval bounds are $l^{(0)}=x-\epsilon$ and $u^{(0)}=x+\epsilon$. For a linear transformation $z=Wh+b$, the interval bounds are propagated as:
\begin{equation}
l_z=
W^{+}l_h+
W^{-}u_h+b,
\end{equation}
\begin{equation}
u_z=
W^{+}u_h+
W^{-}l_h+b,
\end{equation}
where $l_h$ represents the lower bound vector of the input layer, $u_h$ represents the upper bound vector of the input layer, $W^{+}=\max(W,0)$, which keeps only positive weights, and $W^{-}=\min(W,0)$, which only keeps negative weights \cite{gowal2018effectiveness,mirman2018differentiable}. 

Since KANs replace fixed activation functions with learnable spline functions, each KAN layer can be represented as:
\begin{equation}
y_j=
\sum_{i=1}^{n}
\phi_{ij}(x_i),
\end{equation}
where $\phi_{ij}$ denotes a spline activation function. For an input interval $x_i\in[l_i,u_i]$ the lower and upper spline bounds are:
\begin{equation}
\underline{\phi}_{ij}
=
\min_{t\in[l_i,u_i]}
\phi_{ij}(t),
\end{equation}
\begin{equation}
\overline{\phi}_{ij}
=
\max_{t\in[l_i,u_i]}
\phi_{ij}(t).
\end{equation}

The output interval of a KAN neuron is therefore:
\begin{equation}
l_j=
\sum_{i=1}^{n}
\underline{\phi}_{ij},
\end{equation}
\begin{equation}
u_j=
\sum_{i=1}^{n}
\overline{\phi}_{ij}.
\end{equation}

These bounds are recursively propagated through all KAN blocks until the final logit layer. Let $l_k$ and $u_k$ denote the lower and upper bounds of output $k$. For a sample with true class $y$, robustness is certified whenever:
\begin{equation}
l_y
>
\max_{k\neq y}
u_k.
\end{equation}

Under this condition, no perturbation satisfying $\|\delta\|_\infty\le\epsilon$ can alter the classifier prediction \cite{gowal2018effectiveness,mirman2018differentiable}. The certified accuracy under IBP is then calculated as:

\begin{equation}
CA_{IBP}(\epsilon)
=
\frac{1}{N}
\sum_{i=1}^{N}
\mathds{1}
\left(
l_{y_i}
>
\max_{k\neq y_i}
u_k
\right).
\end{equation}
\subsection{Defense Mechanisms}
{In this section, we explain the defense mechanisms that are used during the experiments. These defenses include adversarial training, randomized smoothing, and interval bound propagation.}

\begin{algorithm}[h]
\caption{Adversarial Training.}
\KwIn{Training data: $X_\text{train}$, training labels: $Y_\text{train}$, perturbation bound: $\epsilon$, step size: $\alpha$, iteration number: $I_t$, and number of epochs: $E$.}
model $\gets$ Instantiate(KAN)\;
\For{$e \in \{0,~1,~\dots,~E-1\}$}{
  \For{$(x,~y) \in (X_\text{train},~Y_\text{train})$}{
    $x_{\text{adv}} \gets \text{PGD}(x,~\epsilon,~\alpha,~ I_t)$\;
    $x_\text{mixed} \gets \text{Concatenate}(x,~x_\text{adv})$\;
    $y_\text{mixed} \gets \text{Concatenate}(y,y)$\;
    $y_p \gets \text{model.forward}(x_{\text{mixed}})$\;
    loss $\gets \ell(y_p,~y_\text{mixed})$\;
    model.backward(loss)\;
  }
}
\Return model.\
\label{algorithm:adversarial-training}
\end{algorithm}
Algorithm \ref{algorithm:adversarial-training} shows the adversarial training strategy that we employ during the training session. We use PGD to generate adversarial images within each batch with $8/255$ as perturbation bound, $2/255$ as step size, and $10$ as the number of iterations. Then, we restrict each pixel between $0$ and $1$. After that, we train the model on the mixed batch of benign and adversarial data in order to make the model learn both benign and adversarial distributions.

\begin{algorithm}[h]
\caption{Randomized Smoothing (Training and Inference)}
\KwIn{Training data: $X_\text{train}$, training labels: $Y_\text{train}$, test data: $X_\text{test}$, test labels: $Y_\text{test}$, Gaussian noise level $\sigma$, identity matrix: $\textit{I}$, number of epochs: $E$, and Monte Carlo sample number: $\mathcal{M}$.}
model $\gets$ Instantiate(KAN)\;
\For{$e \in \{0,~1,~\dots,~E-1\}$}{
  \For{$(x,~y) \in (X_\text{train},~Y_\text{train})$}{
  $\xi \gets \mathcal{N}(0,\sigma^2I)$\;
  $x_{\text{noisy}} \gets x + \xi$\;
  $y_p \gets \text{model.forward}(x_{\text{noisy}})$\;
  loss $\gets \ell(y_p,~y)$\;
    model.backward(loss)\;
  }
}
\textbf{Inference phase}:\\
smoothed = SmoothedClassifier(model, $\sigma$, $\mathcal{M}$)\;
standard\_accuracy $\gets$ Evaluate(smoothed, $X_\text{test}$, $Y_\text{test}$)\;
$X_\text{Adversarial} \gets$ Attack(model, $X_\text{test}$, $Y_\text{test}$)\; 
robust\_accuracy $\gets$ Evaluate(smoothed, $X_\text{Adversarial}$, $Y_\text{test}$)\;
\label{algorithm:randomized-smoothing}
\end{algorithm}

Algorithm \ref{algorithm:randomized-smoothing} demonstrates how we employ randomized smoothing during the training session and the test phase. During the training session, we sample from a Gaussian noise distribution with noise level of $0.25$. After that, we obtain noisy images by adding the noise to the original images and train the base classifier on the noisy images. During the inference phase, we generate a smoothed classifier with $0.25$ as noise level and $64$ as the number of Monte Carlo samples. Adversarial data (FGSM, PGD, and C\&W) are generated based on the gradients of the base model; however, the final evaluation is conducted on the smoothed model.

\begin{algorithm}[h]
\caption{Interval Bound Propagation Training.}
\KwIn{Training data: $X_\text{train}$, training labels: $Y_\text{train}$, maximum trade-off coefficient: $\lambda_\text{max}$, number of epochs: $E$, warm-up epochs: $W_E$, ramp-up epochs: $R_E$, maximum perturbation size: $\epsilon_\text{max}$, ramp-up function: $\mathcal{S}_t$.}
\For{$e \in \{0,~1,~\dots,~E-1\}$}{
  $t \gets e+1$\;
  \For{$(x,~y) \in (X_\text{train},~Y_\text{train})$}{
  $\lambda_t$ = $\mathcal{S}_t \times \lambda_\text{max}$\;
  $\epsilon_t$ = $\mathcal{S}_t \times \epsilon_\text{max}$\;
  $y_p \gets \text{model.forward}(x)$\;
  {loss $\gets (1-\lambda_t) \times \ell_\text{Clean}(y_p,~y) + \lambda_t \times \ell_\text{IBP}(y_p,~y,~\epsilon_t)$\;}
    model.backward(loss)\;
  }
}
\label{algorithm:interval-bound-propagation}
\end{algorithm}

Algorithm \ref{algorithm:interval-bound-propagation} demonstrates the IBP-based training. Epochs are divided into three intervals: warm-up, ramp-up, and robust training. In warm-up, perturbation size and trade-off parameter $\lambda$ are equal to $0$ in order to normally train the model. After that, in ramp-up, $\lambda$ gradually increases to $0.5$ and $\epsilon$ gradually increases to $4/255$. Finally, in robust training interval the model is trained on the maximum pre-defined values of $\lambda$ and $\epsilon$. The formula for linear ramp-up function $\mathcal{S}_t$ is as follows:
\begin{equation}
\mathcal{S}_t=
\begin{cases}
    0 \quad &t \leq W_E\\
    \frac{t-W_E}{R_E-W_E} \quad &W_E < t \leq R_E\\
    1 \quad &R_E < t\\
\end{cases}
\end{equation}
where $W_E$ is warm-up epochs, $R_E$ is ramp-up epochs, and $t$ is the epoch number. {Moreover, $\ell_\text{IBP}(y_p,~y,~\epsilon_t)$ constructs the initial interval $[x-\epsilon_t,~x+\epsilon_t]$ and propagates these bounds through the network.} The final loss during IBP-based training is calculated based on Eq. (\ref{formula:IBP-Clean-Loss-Tradeoff}).
\subsection{Threat Model}
\label{sec:threat-model}
We assume strong white-box evasion attacks with full knowledge about the models' architecture, parameters, gradients, and defense mechanism. The attackers cannot modify training data or model parameters and are restricted to test-time perturbations within predefined norm-bounded regions.

Attacks include FGSM, PGD, and C\&W. We use $\ell_\infty$-norm FGSM with perturbation sizes ranging from $0.01$ to $0.09$. For PGD, we use three different variants: standard, multi-step, and adaptive PGD. All the PGD attacks use perturbation bound of $8/255$ and step size of $2/255$. 

In standard PGD, the iteration ranges from $10$ to $50$, while other types generate adversarial samples during $100$ iterations. In multi-step PGD, we use three restarts, and for adaptive PGD, we use three \underline{e}xpectation-\underline{o}ver-\underline{t}ransformation (EoT) samples. Moreover, we use $\ell_2$-norm C\&W to further evaluate the robustness of these KAN-based models. Table \ref{tab:attack-parameters} summarizes the specifications and parameters of each attack.
\begin{table*}[h]
    \centering
    \caption{The parameters of each adversarial attack.}
    \resizebox{\linewidth}{!}{
    \begin{tabular}{l l c}
        \hline
        Attack & Parameters & Norm\\
        \hline
         FGSM & $\epsilon \in \{0.01,~0.02,...,0.09\}$ & $\ell_\infty$\\[0.5em]
         
         Standard PGD & $\epsilon = 8/255$, $\alpha = 2/255$, $i\in \{10,~20,~30,~40,~50\}$&$\ell_\infty$\\[0.5em]

         Multi-Step PGD&$\epsilon = 8/255$, $\alpha = 2/255$, $i=100$, $\text{Restart} = 3$&$\ell_\infty$\\[0.5em]

         Adaptive PGD &$\epsilon = 8/255$, $\alpha = 2/255$, $i=100$, $\text{EoT Samples} = 3$&$\ell_\infty$\\[0.5em]

         C\&W&binary-search steps=9, steps = 1000, step size = $1e-2$, confidence = 0&$\ell_2$\\
         \hline
    \end{tabular}}
    \label{tab:attack-parameters}
\end{table*}
\section{Experimental Setups and Results}
\label{sec:experimental-setup-results}
In this section, we explain the experimental setups, certified results, empirical results, and discuss our findings. The empirical results are averaged over five runs for all the models on both datasets.
\subsection{Datasets and Implementation Details}
We use CIFAR-10 and SVHN datasets. Both datasets include three input channels: red, green, and blue. As a result, we use vectors of size three to normalize the data, where each element correspond to one of the channels. We normalize CIFAR-10 using mean vector of $(0.4914,~0.4822,~0.4465)$ and standard deviation vector of $(0.247,~0.243,~0.261)$. 

Moreover, we normalize SVHN with mean vector of $(0.4377,~0.4438,~0.4728)$ and standard deviation vector of $(0.1980,~0.2010,~0.1970)$. We keep the same normalization values during both training and test sessions.

We used a server equipped with two NVIDIA H100 GPUs. For the main libraries, we used numpy \texttt{2.2.4}, torch \texttt{2.7.0}, torchvision \texttt{0.22.0}, and scipy \texttt{1.15.3}.
\subsection{Evaluated Architectures and Metrics}
We use KAN-Mixers \cite{canuto2025kan}, KANICE \cite{ferdaus2024kanice}, and PoolKANNeXt \cite{lv2025poolkannext} architectures to conduct our experiments. We use these architectures since they have demonstrated state-of-the-art performances in CV-based tasks; this enables us to evaluate the accuracy and robustness of these models more fairly across the datasets.

In order to assess the performance of the selected models, we use \underline{s}tandard \underline{a}ccuracy (SA) and \underline{r}obust \underline{a}ccuracy (RA). SA shows the accuracy of the model under benign dataset, while RA measures the accuracy under adversarial samples. By using these two metrics, we can analyze the performance of the models more deeply, and gain valuable insights of the behavior of these models.
\subsection{Hyperparameters}
 KANICE is trained for 200 epochs, with batch size of 128, and learning rate of $1e-3$. KAN-Mixers is trained for 50 epochs, with batch size of 64, and learning rate of $0.0001282$. PoolKANNeXt is trained for 30 epochs, with batch size of 128, and learning rate of $0.0037$. These settings stay the same during standard, adversarial, RS-based, and IBP-based training sessions. For adversarial training, we use perturbation bound $8/255$, step size $2/255$, and 10 iterations across all the architectures. 

For RS, we set Gaussian noise level ($\sigma$) to 0.25. Moreover, the number of Monte Carlo samples ($\mathcal{M}$) is a test-time parameter, which equals to 64 for all models across all datasets. During IBP-based training sessions, maximum perturbation size is $4/255$ and maximum $\lambda$ equals to 0.5 across all the models. The number of warm-up and ramp-up epochs are 10 and 80 for KANICE, 5 and 40 for KAN-Mixers, and 5 and 20 for PoolKANNeXt. Table \ref{tab:training-inference-hyperparameters} summarizes the hyperparameters that are used during the training session.

\begin{table}[h]
    \centering
    \caption{Training hyperparameters.}
    \resizebox{\linewidth}{!}{
    \begin{tabular}{l | c c c | c c c | c c | c c c c}
        \hline
        Model & \multicolumn{3}{c}{Common} & \multicolumn{3}{c}{AT} & \multicolumn{2}{c}{RS} & \multicolumn{4}{c}{IBP}\\
        \hline
        &epochs&bs&lr&$\epsilon$&$\alpha$&Iterations&$\sigma$&$\mathcal{M}$&$\epsilon_\text{max}$&$W_E$&$R_E$&$\lambda_\text{max}$\\
        \hline
        KANICE&200&128&$1e-3$&$8/255$&$2/255$&10&0.25&64&$4/255$&10&80&0.5\\
        
        KAN-Mixers&50&64&$0.0001282$&$8/255$&$2/255$&10&0.25&64&$4/255$&5&40&0.5\\
        
        PoolKANNeXt&30&128&$0.0037$&$8/255$&$2/255$&10&0.25&64&$4/255$&5&20&0.5\\
        \hline
    \end{tabular}}
    \label{tab:training-inference-hyperparameters}
\end{table} 
\subsection{Certified Robustness Results}
{In this section, we explain the certified robustness results of randomized smoothing.}

\begin{table}[h]
    \centering
    \caption{$\ell_2$-certified robustness of different KAN architectures under randomized smoothing at certification thresholds of 2/255, 4/255, and 8/255 - certified accuracy (CA) and mean radius.}
    \resizebox{0.75\linewidth}{!}{
    \begin{tabular}{l c c c c c}
        \hline
        Model & Dataset & CA@$\frac{2}{255}$ & CA@$\frac{4}{255}$ & CA@$\frac{8}{255}$ & MR\\[0.75ex]
        \hline
        KANICE &CIFAR-10&60.85&60.19&59.01&0.46\\
        
        KAN-Mixers&CIFAR-10&66.84&66.27&65.46&0.75\\
        
        PoolKANNeXt&CIFAR-10&80.33&80.10&79.78&1.18\\
        
        KANICE &SVHN&92.70&92.45&91.93&0.91\\
        
        KAN-Mixers&SVHN&91.38&90.92&90.03&0.71\\
        
        PoolKANNeXt&SVHN&92.36&92.24&92.05&1.10\\
        \hline
    \end{tabular}}
    \label{tab:ca-rs}
\end{table}

Table \ref{tab:ca-rs} shows the $\ell_2$-certified robustness of different KAN architectures under randomized smoothing. \underline{C}ertified \underline{a}ccuracy (CA) measures the percentage of test samples satisfying $R_i \geq r$, where $R_i$ denotes the certified $\ell_2$ radius and $r$ is the certification threshold. The \underline{m}ean \underline{r}adius (MR) represents the average certified radius across all test samples, where larger values indicate stronger certified robustness guarantees. In CIFAR-10, PoolKANNeXt demonstrates the strongest certified robustness, while in SVHN, PoolKANNeXt and KANICE demonstrate competitive certified robustness.
\subsection{Empirical Robustness Results}
Table \ref{tab:sa-fgsm} shows the SA and RA under FGSM attacks for all evaluated architectures and defense strategies on CIFAR-10 and SVHN. Across both datasets, the results reveal a robustness hierarchy, in which adversarial training consistently provides the strongest protection against gradient-based attacks.

\begin{table}[h]
    \centering
    \caption{Standard accuracy and robust accuracy of different KAN models under FGSM attacks.}
    \resizebox{\linewidth}{!}{
    \begin{tabular}{l | c | c c c c c c c c c}
        \hline
        \multirow{2}{*}{\diagbox{Model}{Metric}} & \multirow{2}{*}{SA} & \multicolumn{9}{c}{RA - FGSM}\\

        &&0.01&0.02&0.03&0.04&0.05&0.06&0.07&0.08&0.09\\
        
        \hline
        \multicolumn{11}{>{\columncolor{green!30}}c}{CIFAR-10}\\
        \hline
        KAN-Mixers-Plain&76.17&32.03&28.62&25.64&23.12&21.16&19.34&17.58&16.28&14.92\\ 
        
        KANICE-Plain&85.09&40.74&38.35&36.42&34.87&33.37&32.05&30.60&29.36&28.18\\
        
        PoolKANNeXt-Plain&85.82&36.41&25.73&18.96&14.58&12.12&10.31&9.08&8.03&7.54\\
        
        \hline
        KAN-Mixers-AT&74.42&60.09&57.53&54.85&52.25&49.46&47.04&44.54&42.05&39.90\\
        
        KANICE-AT&84.55&65.74&64.45&63.28&\color{blue}\textbf{61.89}&\color{blue}\textbf{60.48}&\color{blue}\textbf{59.09}&\color{blue}\textbf{58.00}&\color{blue}\textbf{56.96}&\color{blue}\textbf{56.22}\\
        
        PoolKANNeXt-AT&85.27&\color{blue}\textbf{70.21}&\color{blue}\textbf{67.30}&\color{blue}\textbf{64.04}&60.79&57.75&54.51&51.04&48.23&45.56\\
        
        \hline
        KAN-Mixers-RS&73.77&34.21&32.81&31.06&29.50&28.27&26.66&25.23&24.31&22.93\\
        
        KANICE-RS&83.11&46.38&44.92&43.53&42.08&40.65&39.45&37.97&36.66&35.41\\
        
        PoolKANNeXt-RS&\color{blue}\textbf{86.37}&41.90&35.62&30.18&25.54&21.74&18.50&16.40&14.70&13.09\\
        
        \hline
        KAN-Mixers-IBP&75.88&32.31&29.37&26.96&24.94&23.11&21.76&20.20&19.05&18.05\\
        
        KANICE-IBP&\color{blue}\textbf{86.65}&44.35&42.19&40.24&38.44&36.63&35.36&34.06&33.01&31.85\\
        
        PoolKANNeXt-IBP&77.05&35.28&31.79&28.54&25.14&22.70&20.50&17.99&16.20&14.75\\
        \hline
        \multicolumn{11}{>{\columncolor{orange!30}}c}{SVHN}\\
        \hline
        KAN-Mixers-Plain&94.51&62.18&54.54&48.19&42.74&38.31&34.55&31.63&29.03&26.83\\ 
        
        KANICE-Plain&93.39&65.34&63.88&62.35&61.06&59.78&58.74&57.73&56.78&55.94\\
        
        PoolKANNeXt-Plain&92.62&49.71&37.06&28.78&23.26&19.34&16.55&14.50&13.04&11.84\\
        
        \hline
        KAN-Mixers-AT&\color{blue}\textbf{95.74}&\color{blue}\textbf{78.75}&\color{blue}\textbf{77.69}&\color{blue}\textbf{76.49}&\color{blue}\textbf{75.09}&\color{blue}\textbf{73.69}&\color{blue}\textbf{72.26}&\color{blue}\textbf{70.82}&\color{blue}\textbf{69.35}&\color{blue}\textbf{67.76}\\
        
        KANICE-AT&93.90&73.76&72.36&71.07&69.63&68.29&66.93&65.65&64.44&63.26\\
        
        PoolKANNeXt-AT&92.75&73.24&71.48&69.45&67.39&65.17&62.85&60.48&58.09&55.76\\
        
        \hline
        KAN-Mixers-RS&93.78&68.02&65.87&63.29&60.53&57.78&54.58&52.03&49.10&46.64\\
        
        KANICE-RS&91.66&63.51&62.43&61.15&59.59&58.10&56.54&55.02&53.57&52.03\\
        
        PoolKANNeXt-RS&92.57&63.54&59.35&54.36&49.90&45.62&41.82&38.52&35.58&33.03\\
        
        \hline
        KAN-Mixers-IBP&89.07&60.42&55.51&50.79&46.33&42.58&39.21&36.24&33.77&31.45\\
        
        KANICE-IBP&92.34&62.07&60.11&58.05&56.14&54.34&52.76&51.39&50.34&49.19\\
        
        PoolKANNeXt-IBP&90.58&56.65&49.06&42.39&36.76&32.19&28.34&25.05&22.38&20.19\\
        \hline
        
    \end{tabular}}
    \label{tab:sa-fgsm}
\end{table}

On CIFAR-10, clean accuracy varies substantially across architectures. The highest SA is achieved by KANICE-IBP ($86.65\%$) and PoolKANNeXt-RS ($86.37\%$), demonstrating that certified defenses do not necessarily degrade benign performance. However, robustness under FGSM tells a different story. As the perturbation size increases from $0.01$ to $0.09$, all models experience performance degradation. Moreover, adversarially trained models exhibit significantly slower degradation rates. For example, PoolKANNeXt-AT maintains $45.56\%$ RA at $\epsilon = 0.09$, whereas its plain counterpart drops significantly to $7.54\%$. Similarly, KANICE-AT retains $56.22\%$ RA at the strongest perturbation level, substantially exceeding the robustness achieved by KANICE-RS ($35.41\%$) and  KANICE-IBP ($31.85\%$). These observations suggest that exposure to adversarial examples during the training session enables the models to learn decision boundaries that are substantially more resistant to first-order gradient attacks.

\begin{figure}[h]
    \centering
    \includegraphics[width=\linewidth]{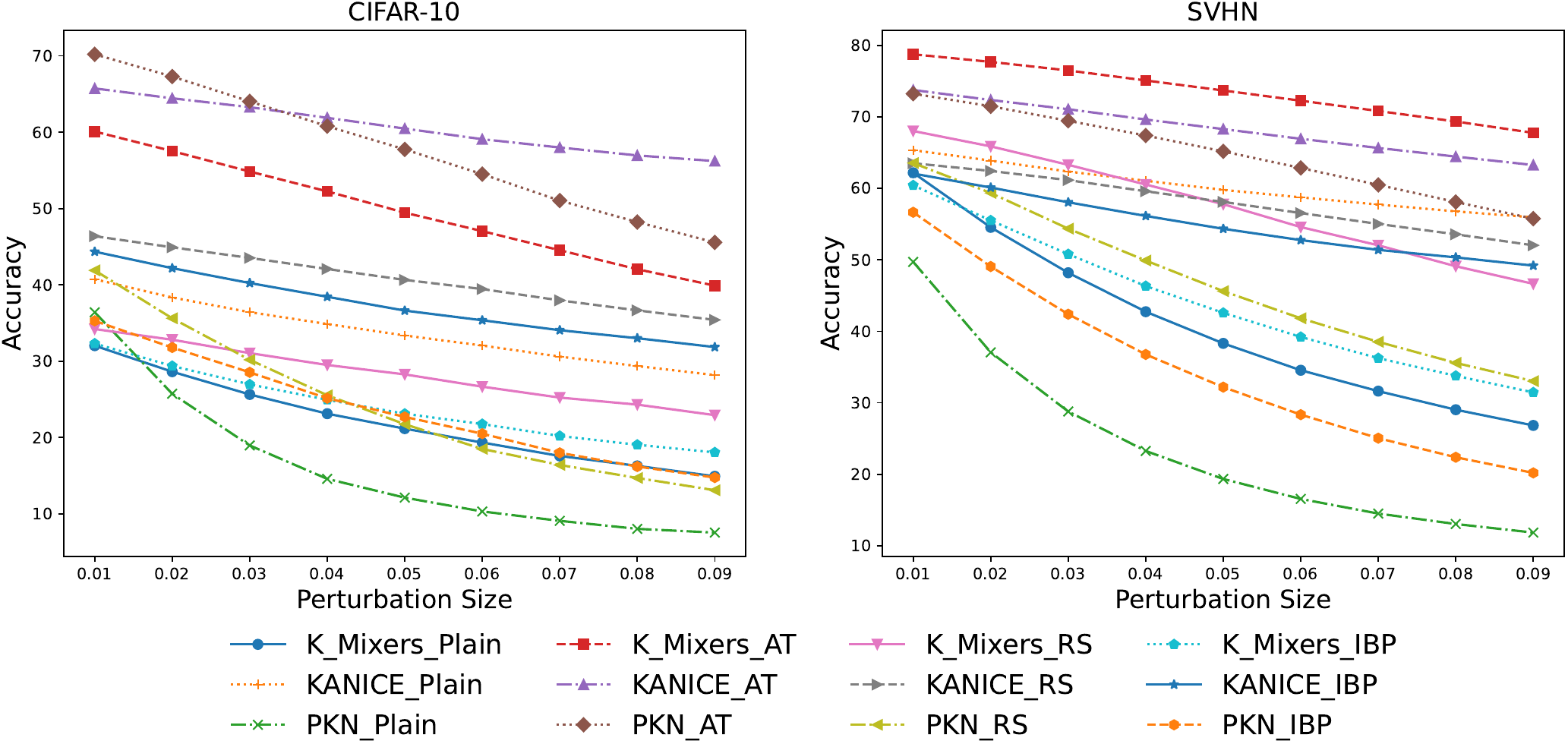}
    \caption{Robust accuracy of KAN models against FGSM across different perturbation sizes on the CIFAR-10 and SVHN datasets.}
    \label{fig:FGSM-results}
\end{figure}

A similar trend is observed on SVHN, although the overall robustness levels are noticeably higher. Nearly all models achieve more than $90\%$ SA, indicating that SVHN is intrinsically easier for the evaluated architectures than CIFAR-10. Furthermore, adversarially trained variants not only preserve robustness but also achieve some of the highest clean accuracies. KAN-Mixers-AT attains $95.74\%$ SA while maintaining $67.76\%$ RA at $\epsilon = 0.09$, outperforming all other defense mechanisms. {Figure \ref{fig:FGSM-results} demonstrate the RA of the models under FGSM with different perturbation sizes. RA consistently decreases as the perturbation size increases, suggesting the more profound effects of the stronger attack configurations. Compared with CIFAR-10, SVHN generally exhibits a steeper decline in RA as perturbation size increases, although its absolute robustness remains consistently higher across perturbation sizes.}

Another notable observation is the limited effectiveness of randomized smoothing and interval bound propagation against FGSM attacks. While both defenses generally improve robustness compared to the undefended models, their gains remain modest relative to adversarial training. This outcome is expected because RS and IBP are designed primarily to provide certified guarantees within predefined perturbation regions rather than explicitly optimizing robustness against the strongest empirical adversarial examples. As a result, their empirical robustness remains lower than that of adversarially trained models despite occasionally achieving higher clean accuracies.

Table \ref{tab:pgd-cw} reports the robust accuracy under multiple PGD variants and the $\ell_2$-based C\&W attack. Compared with FGSM, PGD represents a substantially stronger adversary because it performs iterative optimization to find more effective perturbations. As a result, the robustness gap between defense mechanisms becomes more obvious.

\begin{table}[h]
    \centering
    \caption{Robust accuracy of different KAN models under PGD and C\&W attacks including multi-step (MS) and adaptive (A) PGD.}
    \resizebox{0.9\linewidth}{!}{
    \begin{tabular}{l | c c c c c c c | c}
        \hline
        \multirow{2}{*}{\diagbox{Model}{Metric}} & \multicolumn{7}{c|}{PGD}&C\&W\\

        &10&20&30&40&50&MS&A&\\
        \hline
        \multicolumn{9}{>{\columncolor{green!30}}c}{CIFAR-10}\\
        \hline
         KAN-Mixers-Plain&7.71&7.61&7.60&7.60&7.60&7.59&7.60&24.51\\   
        KANICE-Plain&7.10&6.42&6.39&6.36&6.31&6.08&6.26&27.82\\
        PoolKANNeXt-Plain&2.56&2.39&2.37&2.37&2.36&2.29&2.34&26.91\\
        
        \hline
        KAN-Mixers-AT&41.52&41.42&41.39&41.38&41.38&41.37&41.38&25.48\\
        
        KANICE-AT&38.66&38.10&38.04&38.01&37.99&37.32&38.00&20.71\\
        
        PoolKANNeXt-AT&\color{blue}\textbf{54.77}&\color{blue}\textbf{54.65}&\color{blue}\textbf{54.62}&\color{blue}\textbf{54.61}&\color{blue}\textbf{54.61}&\color{blue}\textbf{54.59}&\color{blue}\textbf{54.60}&\color{blue}\textbf{31.17}\\
        
        \hline
        KAN-Mixers-RS&25.20&25.18&25.38&25.37&25.41&25.42&25.47&25.52\\
        
        KANICE-RS&34.59&34.46&34.65&34.88&34.83&34.89&34.65&20.58\\
        
        PoolKANNeXt-RS&11.80&11.55&11.50&11.51&11.52&11.43&11.38&23.51\\
        
        \hline
        KAN-Mixers-IBP&9.19&9.08&9.06&9.05&9.05&8.99&9.05&22.42\\
        
        KANICE-IBP&12.44&11.88&11.82&11.81&11.81&11.62&11.80&26.85\\
        
        PoolKANNeXt-IBP&16.57&16.34&16.29&16.29&16.29&16.29&16.30&22.13\\
        \hline
        \multicolumn{9}{>{\columncolor{orange!30}}c}{SVHN}\\
        
        \hline
        KAN-Mixers-Plain&25.81&25.04&24.95&24.90&24.89&24.75&24.87&72.09\\   
        KANICE-Plain&39.34&37.98&37.90&37.89&37.88&37.59&37.86&66.15\\
        
        PoolKANNeXt-Plain&12.38&11.99&11.90&11.88&11.86&11.85&11.83&60.71\\
        
        \hline
        KAN-Mixers-AT&\color{blue}\textbf{69.10}&\color{blue}\textbf{68.92}&\color{blue}\textbf{68.90}&\color{blue}\textbf{68.90}&\color{blue}\textbf{68.90}&\color{blue}\textbf{68.85}&\color{blue}\textbf{68.90}&61.75\\
        
        KANICE-AT&51.66&51.08&51.03&51.01&51.00&50.81&50.98&61.42\\
        
        PoolKANNeXt-AT&63.15&63.04&63.03&63.03&63.03&63.02&63.03&50.42\\
        
        \hline
        KAN-Mixers-RS&56.28&56.25&56.21&56.10&56.22&56.09&56.18&\color{blue}\textbf{72.58}\\
        
        KANICE-RS&51.41&51.40&51.37&51.29&51.46&51.32&51.46&51.30\\
        
        PoolKANNeXt-RS&43.63&43.44&43.39&43.32&43.30&43.31&43.33&67.47\\
        
        \hline
        KAN-Mixers-IBP&33.50&33.05&32.99&32.96&32.96&32.90&32.95&62.93\\
        
        KANICE-IBP&32.57&31.28&31.23&31.21&31.20&31.02&31.18&60.61\\
        
        PoolKANNeXt-IBP&29.77&29.29&29.21&29.17&29.18&29.11&29.17&44.03\\
        \hline
        
    \end{tabular}}
    \label{tab:pgd-cw}
\end{table}

For CIFAR-10, adversarial training again emerges as the most effective defense. PoolKANNeXt-AT achieves the highest robustness under all PGD configurations, maintaining approximately $54.6\%$ RA regardless of attack iterations or attack variants. KAN-Mixers-AT and KANICE-AT follow with robust accuracies of approximately $41\%$ and $38\%$, respectively. In contrast, the corresponding undefended models achieve less than $8\%$ RA under PGD, highlighting the severe vulnerability of standard KAN architectures to iterative gradient-based attacks. Interestingly, robustness remains relatively stable as the number of PGD iterations increases from $10$ to $50$ and even under multi-step and adaptive PGD variants, indicating that the attacks have largely converged and that the reported robustness values represent meaningful lower-bound estimates.

The performance of randomized smoothing under PGD is mixed. Although RS substantially improves robustness compared with the plain models, it remains consistently inferior to adversarial training. For example, KANICE-RS achieves approximately $35\%$ RA under PGD, while KANICE-AT maintains approximately $38\%$. Similarly, IBP provides only moderate improvements over the baseline models. These findings suggest that certified defenses offer limited empirical protection against strong optimization-based attacks despite providing formal robustness guarantees within specific perturbation bounds.

The SVHN results exhibit the same overall ranking but with considerably higher robustness values. KAN-Mixers-AT achieves the strongest PGD robustness at approximately $69\%$, followed by PoolKANNeXt-AT. Randomized smoothing also performs notably better on SVHN than on CIFAR-10, with KAN-Mixers-RS maintaining approximately $56\%$ robustness across all PGD variants. This behavior further supports the observation that SVHN is inherently more resilient to adversarial perturbations and allows certified defenses to operate more effectively.

The C\&W results reveal a different trend. Unlike FGSM and PGD, where adversarial training dominates, randomized smoothing frequently provides competitive robustness under the optimization-based $\ell_2$ attack. On CIFAR-10, PoolKANNeXt-AT achieves the highest robustness ($31.17\%$), but several RS and plain-model configurations achieve comparable performance. On SVHN, KAN-Mixers-RS attains the highest robustness ($72.58\%$), slightly exceeding both the plain and adversarially trained variants. This indicates that the smoothing mechanism can effectively mitigate perturbations generated in the $\ell_2$ space, making RS particularly suitable against C\&W-style attacks.

\begin{figure}[h]
    \centering
    \includegraphics[width=\linewidth]{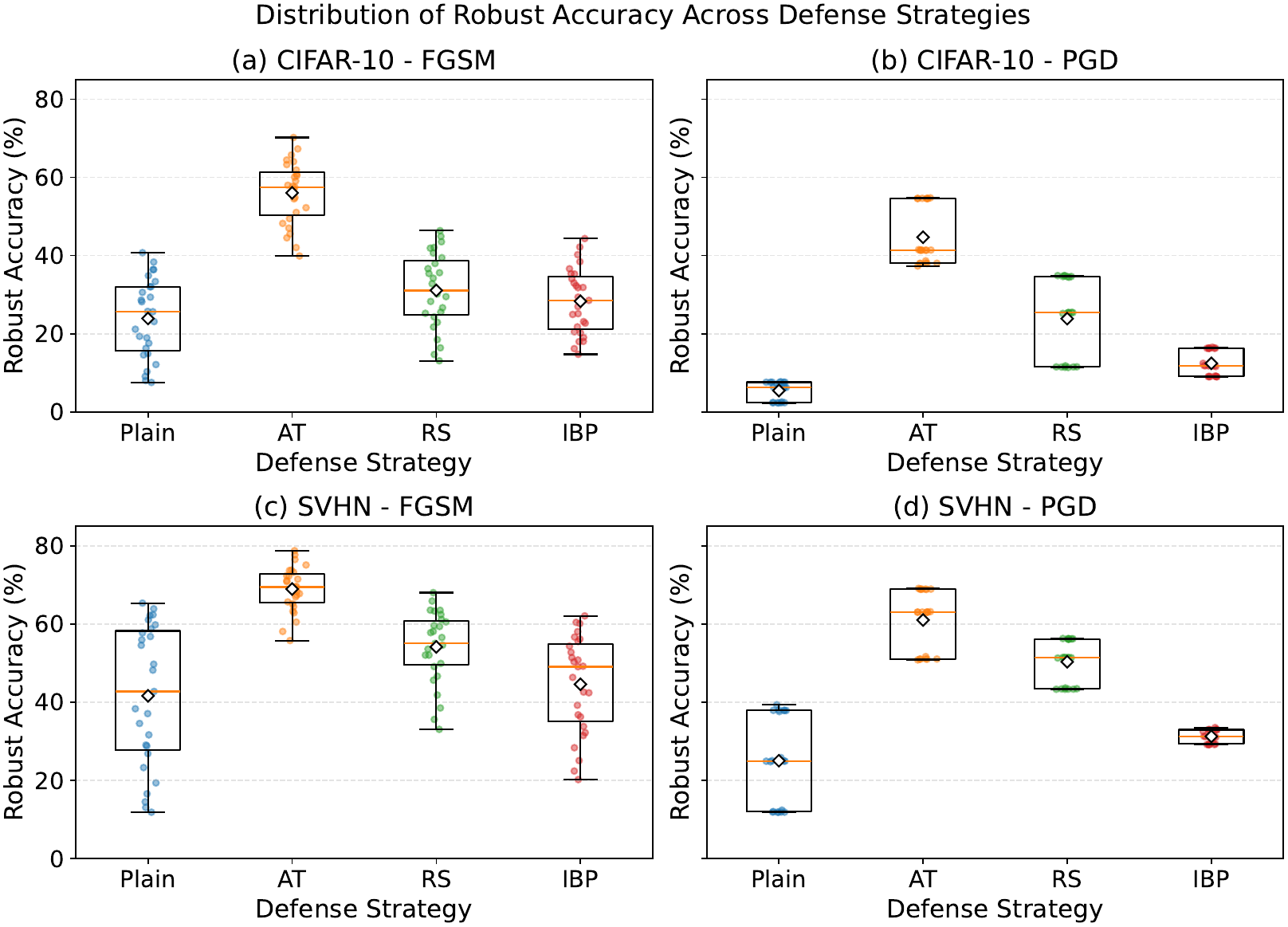}
    \caption{Box plots of robust accuracy across defense strategies under FGSM and PGD attacks on CIFAR-10 and SVHN.}
    \label{fig:boxplot}
\end{figure}

{Figure \ref{fig:boxplot} provides a distributional comparison of RA across the evaluated defense strategies by aggregating the results of all selected KAN architectures over different attack configurations. For FGSM, each box represents the RA values obtained across nine perturbation sizes, while for PGD, each box summarizes the results across the seven evaluated PGD configurations. The figure reveals a clear separation between AT and the other defense strategies, with AT demonstrating the RA distributions upward on both CIFAR-10 and SVHN. This confirms that AT provides the strongest overall empirical protection against FGSM and PGD attacks. The advantage of AT is particularly evident under PGD, where the plain models exhibit very low robust accuracy on CIFAR-10, while the adversarially trained variants maintain substantially higher values. RS generally provides the second strongest protection, particularly on SVHN, while IBP produces more moderate empirical robustness despite its certified-training objective. The distributions also indicate that SVHN models generally maintain higher robust accuracy than their CIFAR-10 counterparts across defense strategies. 

Furthermore, the spread of the box plots suggests that robustness remains architecture-dependent. Although a defense may improve the overall robustness distribution, its effectiveness varies among KAN-Mixers, KANICE, and PoolKANNeXt. These findings suggest that architectural choices, including interactive convolutional blocks, pooling-based feature aggregation, and mixer-style token processing, influence both empirical and certified robustness in different ways, despite all models being built upon the same Kolmogorov-Arnold learning framework.}
\section{Conclusion}
\label{sec:conclusion}
In this paper, we evaluated the empirical robustness of different defended and undefended KAN architectures against FGSM, PGD, and C\&W attacks. While adversarial training proved to be the most effective defense strategy against gradient-based attacks, randomized smoothing demonstrated a competitive performance against $\ell_2$-norm C\&W attack. Moreover, the choices of architecture can directly affect both certified and empirical robustness.

Our future research will focus more deeply on certified robustness of KAN models by employing more certification techniques including verification and convex relaxation. Moreover, investigating more diverse architectures of KAN models in terms of certified and empirical robustness is another future direction. By investigating these areas and other diverse set of defenses, the limitations and strengths of these models could be further investigated.
\bibliographystyle{elsarticle-harv}
\bibliography{main}
\end{document}